\documentclass{article}
\usepackage{graphicx} 
\usepackage{url,hyperref} 
\usepackage{geometry}
\usepackage{cite}
\usepackage{amssymb,amsmath,physics}
\usepackage{subcaption}
\usepackage{nicefrac}
\usepackage{tikz}
\usetikzlibrary{calc,patterns,decorations.pathmorphing}
\tikzset{>=latex}

\title{The Weak Form Is Stronger Than You Think}
\author{Daniel A.~Messenger, April Tran, Vanja Dukic, David M.~Bortz\\
Department of Applied Mathematics\\
University of Colorado, Boulder, CO 80309-0526}

\begin{document}

\maketitle

\abstract{The weak form is a ubiquitous, well-studied, and widely-utilized mathematical tool in modern computational and applied mathematics.  In this work we provide a survey of both the history and recent developments for several fields in which the weak form can play a critical role. In particular, we highlight several recent advances in weak form versions of equation learning, parameter estimation, and coarse graining, which offer surprising noise robustness, accuracy, and computational efficiency.

We note that this manuscript is a companion piece to our October 2024 SIAM News article of the same name.  Here we provide more detailed explanations of mathematical developments as well as a more complete list of references.  Lastly, we note that the software with which to reproduce the results in this manuscript is also available on our group’s GitHub website (\url{https://github.com/MathBioCU}). }

\section{Background}

For a broad class of differential equations, the \emph{weak form} is created by convolving both sides with a sufficiently smooth function $\phi$, integrating over a domain of interest $\Omega$, and using integration-by-parts to obtain an equation with fewer derivatives.   

Historically, this idea originated from the fact that physical conservation laws can commonly be cast in integral and variational formulations. Doing so allows for (in many cases) easier analysis and simulation of a broad class of models, including those with shocks and other solution discontinuities. In the early twentieth century, Sobolev was the first to suggest that $\phi$ be (what Friedrichs later named a \emph{mollifier}) a smooth $C^\infty$ function which integrates to one.  Building on this, Argyris, Courant, Friedrichs, Galerkin, Hrennikoff, Oganesya, and many others contributed to literature leading to computational approaches such as the finite element method (FEM) for solving a differential equation \cite{Oden1990Ahistoryofscientificcomputing}. On the theoretical side, Schwartz rigorously recast the classical notion of a function acting on a point to one acting on a measurement structure or \emph{test function} ($\phi$) \cite{Schwartz1950} and Lax and Milgram then built on this (and other) results to prove the existence of weak solutions (in a Hilbert space) to certain classes of parabolic PDEs \cite{LaxMilgram1955ContributionstotheTheoryofPartialDifferentialEquations}.

Thus, the weak form is a ubiquitous, well-studied, and widely-utilized tool in modern computational and applied mathematics. Importantly, the dominant application for which it has been used is in the study of \emph{solutions} to differential equations, both via analytical and computational approaches.  However, there are a multitude of other applications that benefit from leveraging the conversion of an equation to its weak form.  At its core, the act of convolving a test function $\phi$ with data is a (kernel) smoother, filtering noise and providing a locally averaged estimate of the true state of the system.  Yet, the conversion of an equation to the weak 
form is more powerful than just smoothing the data; choosing a test function asserts a topology or scale through which to view the equation and its solutions. This perspective leads to viewing the conversion as a projection of the data onto the solution manifold while implicitly filtering the noise.  Indeed, recent advances suggest that with a data-driven topology (encoded in the form of $\phi$), weak form versions of equation learning, parameter estimation, and coarse graining offer surprising noise robustness, accuracy, and computational efficiency.

In what follows, we survey the history and state-of-the-art for several fields in which the weak form can play a critical role. We then illustrate several intriguing and promising applications, including the discovery of differential equation models from highly noise-corrupted data, robust and fast parameter inference, and coarse-graining of dynamical systems, all of which substantially benefit from a weak form approach.  

Lastly, we note that this manuscript is a companion to our recently published SIAM News article of the same name \cite{MessengerTranDukicEtAl2024SIAMNews}, but with more detailed explanations of historical and mathematical developments as well as a notably more complete list of references.

\section{Learning Governing Equations}
Historically, model creation has been performed by those with mathematical or statistical training, frequently in collaboration with disciplinary scientists. Starting in the 1970's, however, several researchers began developing methods to automate the process for discovering governing models, including attempts to quantify a \emph{best} model among several candidates. In the statistics literature, Akaike \cite{Akaike1974IEEETransAutomControl} was the first to show that the Kullback-Leibler divergence \cite{KullbackLeibler1951AnnMathStatist} between a model and data could be approximated using (what is now called) the Akaike Information Criteria (AIC). This allowed for computation of a measure of the information lost in using a model to represent the data. Among a set of candidate models, the smallest AIC value thus offers the most parsimonious description of the data and the most efficient flow of information to the parameters.  In the computer science and artificial intelligence literature, several efforts in the 1970's and 1980's were made within the framework of \emph{heuristic theory}, e.g., DENDRAL \cite{FeigenbaumBuchananLederberg1971MachineIntelligence}, 
BACON \cite{Langley1977ProcFifthIntJtConfArtifIntellI}, AM \cite{Lenat1977MachineIntelligence}, and EURISKO \cite{Lenat1983ArtifIntell}.  The idea behind these packages was to codify a set of heuristics (i.e., \emph{rules of thumb}) which themselves included heuristics for altering their own rules and properties, with the goal of discovering empirical laws.

Interest in discovery systems waxed and waned over the years until the release of Schmidt and Lipson's \emph{Eureqa} software, which drew significant attention for it's ability to discover physical (and interpretable) mathematical models and their constitutive parameters using a symbolic regression approach \cite{SchmidtLipson2009Science}. Symbolic regression places no restrictions or assumptions on the mathematical form of the governing model, however, this flexibility comes at the computational cost of being an NP-hard problem.

More recently, there has been an explosion of activity based on the Sparse Identification of Nonlinear Dynamics (SINDy) method \cite{BruntonProctorKutz2016ProcNatlAcadSci}. SINDy uses sparse regression to identify weights $\mathbf{w}=\{w_j\}_{j=1}^J$ in an equation error residual 

\begin{equation}
\left\Vert\partial_t\mathbf{U}-\sum_{j=1}^J w_j f_j(\mathbf{U})\right\Vert_2^2
\label{eq:EEResid}
\end{equation}
(for data $\mathbf{U}$). The sparsification of $\mathbf{w}$ prunes the potential equation terms $\{f_j\}_{j=1}^J$ to simultaneously learn the governing model equation and estimate the parameters.  This approach differs substantially from the symbolic regression in that SINDy starts with a library to be trimmed, resulting in models that are limited by the initial library, but yields an overall learning method which is computationally much faster.
 
 Although residual-based methods (such as SINDy) are computationally efficient, the use of noisy data presents a significant challenge due to the need to approximate derivatives of the data (e.g., see the approximations in \cite{RudyBruntonProctorEtAl2017SciAdv}). Building on SINDy, several groups \cite{SchaefferMcCalla2017PhysRevE,GurevichReinboldGrigoriev2019Chaos, MessengerBortz2021MultiscaleModelSimul, MessengerBortz2021JComputPhys, PantazisTsamardinos2019Bioinformatics, WangHuanGarikipati2019ComputMethodsApplMechEng}, 
 independently discovered that learning using the weak form of the model bypassed both the approximation question \emph{and} was highly robust to noise. The core of this idea is that by multiplying both sides with a compactly supported test function $\phi\in C^{\infty}_{\mathrm{c}}(\Omega)$ allows repeated integration-by-parts to move all derivatives from the state variables onto the test function. To illustrate, if we consider a feature library consisting of spatial derivatives up to order $K$ acting on polynomials up to order $P$, the (weak form) residual is \[\left\Vert\left<\partial_t\phi,\mathbf{U}\right>+\sum_{k=0}^{K}\sum_{p=0}^{P}(-1)^k w_{k,p}\left<\partial_x^k\phi, \mathbf{U}^p\right>\right\Vert_2^2.\] Naturally, relocation of all derivatives to the test function is only possible when the equation can be written as a differential operator acting on a linear combination of nonlinear functions of the data. However, this does encompass an incredibly broad set of models across the sciences and engineering;  Kuramoto-Sivashinsky, Nonlinear Schrödinger, Lorenz, FitzHugh-Nagumo, and many others all fall into this category.\footnote{We note that researchers have developed strategies (under certain conditions) to bypass this limitation, see the comment in Section \ref{sec:ParEst} about \cite{Pearson1992Proc31stIEEEConfDecisControl}.} Indeed, other researcher have also found success using weak/variational form versions of SINDy to accurately recover models for turbulent fluid flow \cite{ReinboldKageorgeSchatzEtAl2021NatCommun,GurevichReinboldGrigoriev2022arXiv210500048}, active matter \cite{GoldenGrigorievNambisanEtAl2023SciAdv}, pattern-forming reaction-diffusion systems (Schnakenberg kinetics, Cahn-Hilliard, Allen-Cahn, etc.) \cite{WangHuanGarikipati2019ComputMethodsApplMechEng,Garikipati2024Data-drivenModellingandScientificMachineLearninginContinuumPhysics,SrivastavaGarikipati2024IntJNumerMethodsEng,WangWuGarikipatiEtAl2020TheorApplMechLett,WangHuanGarikipati2021ComputMethodsApplMechEng}, cancer cell migration \cite{KinnunenSrivastavaWangEtAl2023arXiv230209445}, hybrid systems in ecology \cite{MessengerDwyerDukic2024arXiv240520591}, computational plasma physics \cite{VaseyMessengerBortzEtAl2023arXiv231205339}, and hydrodynamic equations in quantum systems \cite{KharkovShtankoSeifEtAl2021arXiv211102385}.
 
\begin{figure}

\includegraphics[clip,trim={20 110 70 85},width=1\textwidth]{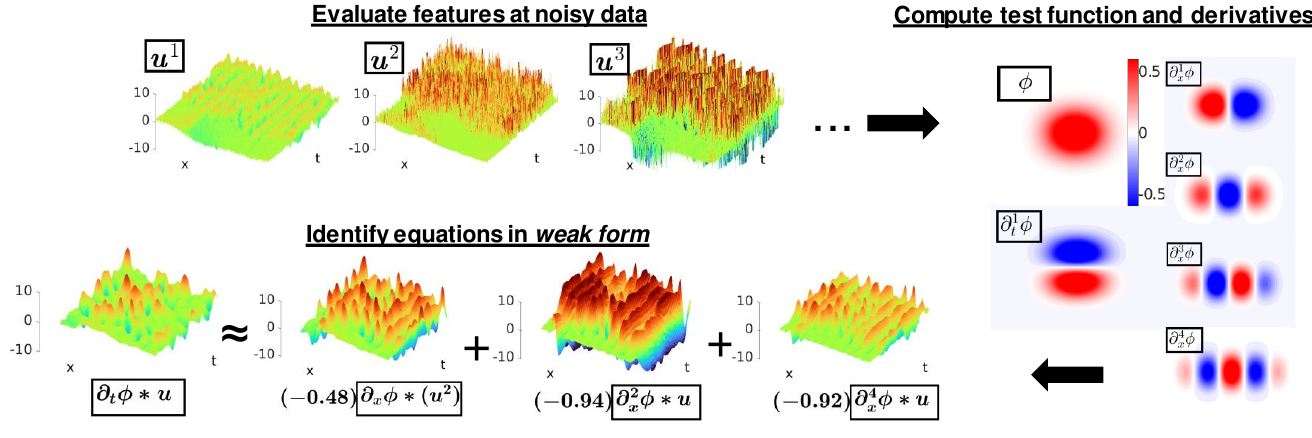}
\caption{\label{WSINDy_diagram}{\bf Schematic of weak-form PDE identification using the \texttt{WSINDy\_PDE} algorithm.} Solution data from the Kuramoto-Sivashinsky(KS) equation with 50\% added noise is collected (z-axis limited to $[-10,10]$ for clarity). From noisy feature evaluations, a reference test function $\phi$ is identified to balance noise filtering and accuracy. Weak-form features are constructed using convolutions against $\phi$ and its derivatives. The governing equations approximately hold in this weak-form space, allowing accurate identification of model terms and coefficients.}

\end{figure}

Figure \ref{WSINDy_diagram} is a flowchart detailing the weak form equation learning framework applied to discovering the Kuramoto-Sivashinsky (KS) PDE in the presence of 50\% additive i.i.d.~Gaussian measurement noise ($\approx3$ dB SNR). In this example, the candidate library consists of all unique operators $u\mapsto \partial_x^k(u^p)$ for $0\leq k,p\leq 6$, i.e., 43 terms (including the true 3-term model)
 .\footnote{This creates a library of 43 terms, comprising derivatives in space acting on powers of the solution $u$. We have bounded $k$ and $p$ at 6, although the creation of larger libraries is, of course, possible.
 In practice, reasonable upper bounds for $k$ and $p$ are frequently known, and thus the ability to discover terms with larger $k$ or $p$ 
 is of questionable value.
 } A mathematically justified choice for the test function $\phi$ is critical to performance; in this example, we match the spectral properties of the test functions to that of the data (as proposed in our weak SINDy (WSINDy) PDE article \cite{MessengerBortz2021JComputPhys}) to filter high frequency noise and preserve the solution signal. Centering shifted copies of test functions at each sample point creates a regression problem for the coefficients $\mathbf{w}$ and allows for accurate PDE discovery from noisy data in under a second on a standard laptop. This performance of this weak form approach is in direct contrast to strong form methods, e.g., data with anything more than 1\% noise will cause SINDy to fail to learn the Navier-Stokes equation.

\begin{figure}
\begin{center}
\includegraphics[clip,trim={0 0 0 0},height=1.85in]{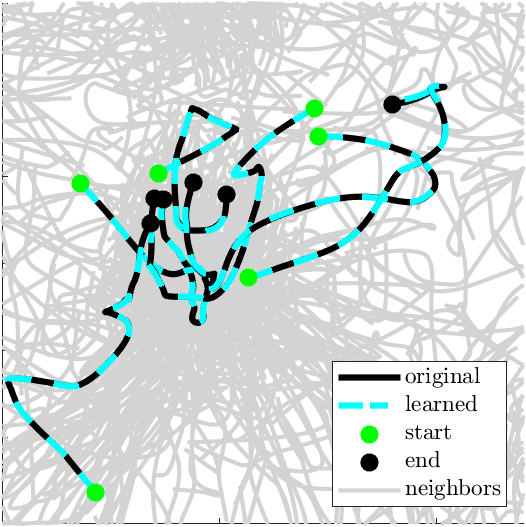}
\hfill
\begin{tikzpicture}
    \node(a){\includegraphics[height=1.85in]{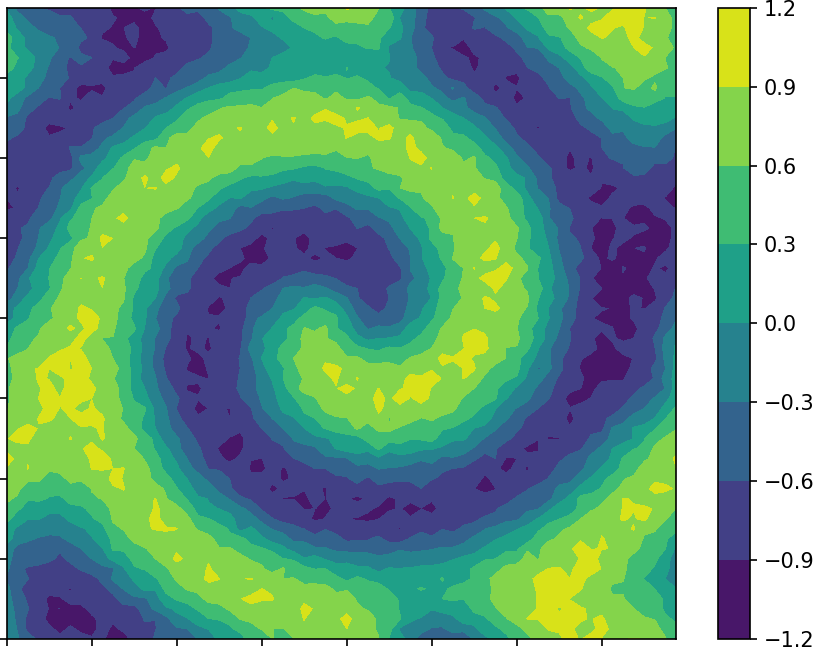}};
    \node at (a.south east)
    [
    anchor=center,
    xshift=-24mm,
    yshift=12mm
    ]
    {
        \includegraphics[height=0.85in]{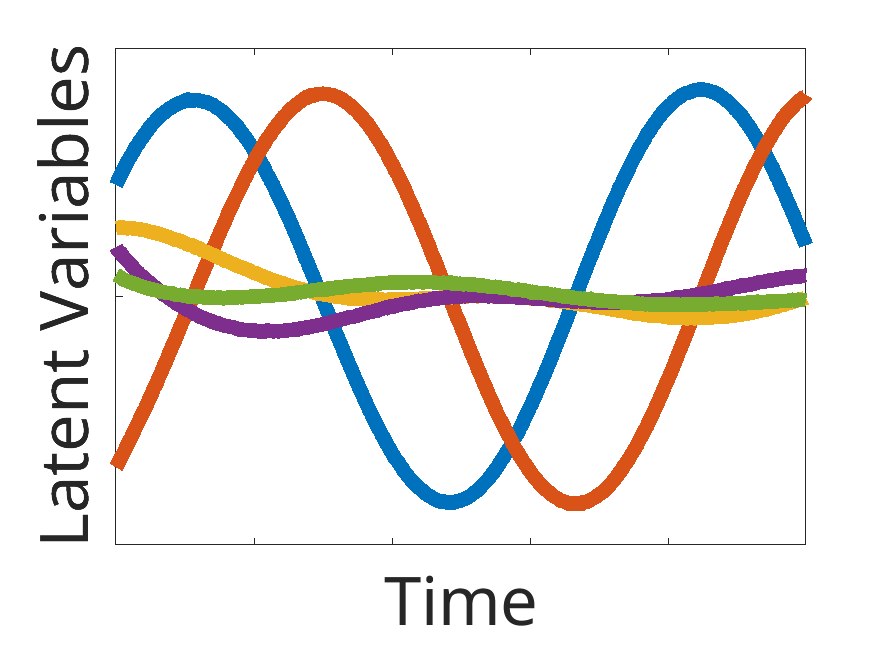}
    };
\end{tikzpicture}
\hfill\includegraphics[clip,trim={0 0 0 0},height=1.85in]{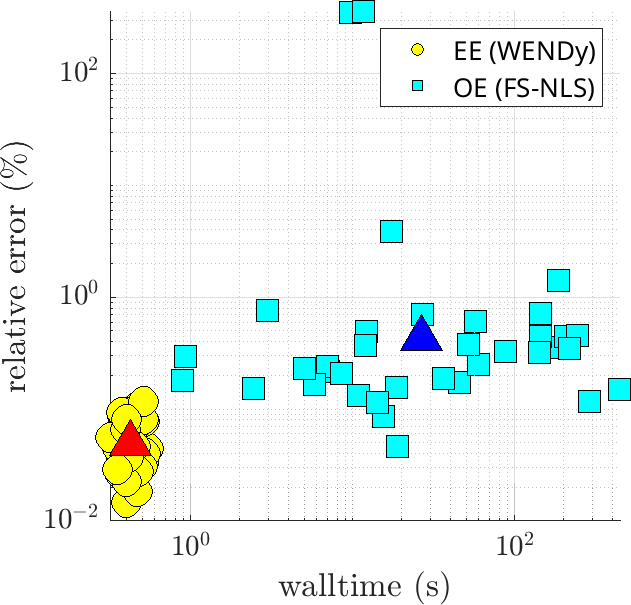}
\end{center}
\caption{\label{WSINDy_JRSII} 
{\bf Equation Learning and Parameter Estimation.} Left: For unlabeled particle trajectories (gray) in a multi-species population, force potentials are learned for each particle and then sorted into species (teal trajectories share a common learned model)  (see \cite{MessengerWheelerLiuEtAl2022JRSocInterface}). Center: Contour snapshot of the noisy measurements of the activator in a Reaction-Diffusion System with (inset) 5D ROM latent space (see \cite{TranHeMessengerEtAl2024ComputMethodsApplMechEng}). Right: Comparison of parameter estimation performance on KS using equation error (EE) and output error (OE) methods (see \cite{BortzMessengerDukic2023BullMathBiol}). Yellow circles represent WENDy, teal squares represent forward solver-NLS, and triangles are the corresponding geometric means.
}

\end{figure}

 Furthermore, the discovery capabilities of the weak form are broader than just finding a canonical ODE or PDE to describe the data (as is done in \cite{MessengerBortz2021JComputPhys,WangHuanGarikipati2019ComputMethodsApplMechEng,PantazisTsamardinos2019Bioinformatics,GurevichReinboldGrigoriev2019Chaos,MessengerBortz2021MultiscaleModelSimul}). 
For example, asymmetric force potentials modeling attraction/repulsion, alignment, and drag can be learned for \emph{each particle} in a deterministic interacting particle system (IPS) model of collective motion (see below for stochastic IPSs). In Figure \ref{WSINDy_JRSII} (left), the (gray) unlabeled trajectories illustrate the motion of a heterogeneous population where subsets of particles are governed by a common learned force model. A weak form method (in this case WSINDy) can rapidly (and parallelizably) learn particle-specific potentials (in less than $10$ seconds) that lead to accurate trajectory predictions. Moreover, models can be clustered to discover the population structure (the teal curves are from one sub-population), offering for example, a novel tool for biologists to study cell population heterogeneity from movement trajectories \cite{MessengerWheelerLiuEtAl2022JRSocInterface}.

The weak form can also augment existing methods, such as in the creation of a reduced order model (ROM) from noise-corrupted or stochastic data by robustly learning governing equations for variables in a latent space. For example, Fries et al., \cite{FriesHeChoi2022ComputMethodsApplMechEng} proposed the Latent Space Dynamics Identification (LaSDI) method, which is a ROM that uses an autoencoder to discover the latent space variables and SINDy to learn the latent space dynamics (similar to the strategy in \cite{ChampionLuschKutzEtAl2019ProcNatlAcadSci}).  While it is effective for learning a ROM that can be hundreds of times faster than a full model simulation, noise corrupted training data will result in an incorrect ROM.  LaSDI can be extended to create the Weak form LasDI (WLaSDI) in which the latent dynamics are learned in the weak form, making the ROM robust to noise in the training data \cite{TranHeMessengerEtAl2024ComputMethodsApplMechEng}. Figure \ref{WSINDy_JRSII} 
 (center) illustrates the results of applying WLaSDI to noisy measurements of the solution to a Reaction-Diffusion system, yielding a ROM with 200 times speedup and around $4\%$ solution error (with the same data, a LaSDI ROM has over 100\% solution error). Even when the noise level is increased to $100\%$, WLaSDI still returns a ROM with under $10\%$ relative error, while a LaSDI ROM would have over 200\% error.  We direct the interested reader to \cite{TranHeMessengerEtAl2024ComputMethodsApplMechEng} for more examples.  We also note the recent extension (WgLaSDI \cite{HeTranBortzEtAl2024arXiv240700337}) which uses a greedy algorithm to simultaneously learn both the latent space dynamics and the autoencoder weights, effectively merging WLaSDI \cite{TranHeMessengerEtAl2024ComputMethodsApplMechEng} with greedy LaSDI (gLaSDI) \cite{HeChoiFriesEtAl2023JournalofComputationalPhysics} to enhance its robustness to noisy data.


\section{Parameter Estimation}\label{sec:ParEst}

At the core of SINDy-type methods lies an \emph{equation error} (EE) regression problem in which the data is substituted directly into the equation and parameters are estimated by minimizing the equation residual \eqref{eq:EEResid}.  This approach is in contrast with \emph{output error} (OE) methods for which the equation is first solved using candidate parameter values and then the resulting approximate solution is compared with data, leading to iterated revision of the parameter estimate until the approximate solution matches the data.

The natural idea of using an equation residual in system identification has been studied extensively over the years.\footnote{To the best of the authors' knowledge, the first designation of an equation residual as an \emph{Equation Error} problem was in the early 1960's by Potts, Ornstein, and Clymer \cite{PottsOrnsteinClymer1961PapPresentMay9-111961WestJtIRE-AIEE-ACMComputConf-IRE-AIEE-ACM61West}.} Using regression with EEs to perform parameter estimation can be traced back to the middle of the 20th century at the dawn of both computers and modern control theory (see \cite{Young1981Automatica} for a good review of the significant results up until 1980).  Much of the early work appeared in the 1950's and 1960's in the aerospace literature \cite{Greenberg1951NACATN2340,BriggsJones1953NACATN2977,Lion1967AIAAJournal} due to the need for rapid system identification in aircraft control engineering.  The methods at that time can be broadly categorized into OE, EE and FE (Fourier Error) approaches, where the FE methods involve converting the DE to an equation in the complex domain and leading to the widely used transfer function methods \cite{SteiglitzMcBride1965IEEETransAutomatContr,Ljung1999}. While OE and FE methods have been more prevalent, EE-based system identification methods continued to be extensively developed in both the control literature (see \cite{AstromEykhoff1971Automatica,GoodwinPayne1977,UnbehauenPrasadaRao1987,Ljung1999,KleinMorelli2006} for broad reviews of system identification, including EE approaches) as well as other fields \cite{Varah1982SIAMJSciandStatComput,PoytonVarziriMcAuleyEtAl2006ComputChemEng,RamsayHookerCampbellEtAl2007JRStatSocSerBStatMethodol,BrunelClairon2015ElectronJStatist,LiangWu2008JAmStatAssoc,DingWu2014StatSin,WangZhou2021IntJUncertainQuantif,Bellman1969MathBiosci}

To directly deal with the challenge of needing derivatives of (potentially noisy) data in EE methods, Shinbrot was the first to propose using a convolution of the model system with compactly supported test functions $\phi$ \cite{Shinbrot1954NACATN3288,Shinbrot1957TransAmSocMechEng}, thus recasting system identification as an algebra problem. In the case where the model is linear in the parameters, the problem is now one of linear regression. This approach was independently rediscovered in the 1960's by Loeb and Cahen \cite{LoebCahen1965Automatisme,LoebCahen1965IEEETransAutomControl} who named it the \emph{Modulating Function Method} (MFM).  Thus, weak form system identification in the control theory and engineering literature fell under the MFM umbrella, while there was no unifying description in other fields.\footnote{For examples of weak form and proto-weak form methods in the applied math and statistics literature, see \cite{BrunelClairondAlche-Buc2014JAmStatAssoc,LiangWu2008JAmStatAssoc,LiuChangChen2016NonlinearDyna}.}  As the MFM method has experienced (more or less) sustained research attention over the last 70 years, this section is notably longer than other sections in this manuscript.

It is well known that it can be challenging to compute a pointwise derivative for noisy data. In an effort to gain computational accuracy and robustness against noise, researchers have proposed several different classes of \emph{modulating functions} (i.e., test functions). Shinbrot's original $\phi$ functions were based on powers of $\sin(\omega t)$,
with additional restrictions on $\omega$ and a piecewise definition near one boundary of the compact support domain \cite{Shinbrot1957TransAmSocMechEng}. Multiple authors \cite{Strejc1961ActaTech,JosephLewisTou1961TransAIEEPartIIApplicatInd} have also developed a framework around repeated integration of measured data, i.e., repeated convolutions with constant $\phi$'s of different support widths.\footnote{We note that the benefit of the integral form for EE system ID has also been rediscovered many times over the years from Diamessis in 1965 \cite{Diamessis1965ProcIEEE} all the way up to contemporary works in the last few years \cite{SchaefferMcCalla2017PhysRevE,VujacicDattner2018StatProbabLett,YaariDattner2019JOSS} among many others.} Loeb and Cahen were inspired by Schwartz functions \cite{Schwartz1950} and proposed a $\phi$ with zeros at all sample points as well as zero derivatives on a subset of points, i.e., a spline test function \cite{LoebCahen1965Automatisme,LoebCahen1965IEEETransAutomControl}. Takaya developed the function
\[
\phi^{(n)}(r) =e^{-\left(\frac{r^2}{2}\right)}\frac{H_n(r)}{\sqrt{2\pi}}
\] for the $n$th derivative of $\phi$ with radius $r$ and where $H_n$ the $n$th Hermite interpolating polynomial \cite{Takaya1968IEEETransAutomControl}.  Perdreauville and Goodson \cite{PerdreauvilleGoodson1966JBasicEng}  were the first to discuss the value in constructing $\phi$ as a multiplicatively separable product of (potentially different) test functions in each independent variable, e.g., $\phi(t,x)=\phi_1(t)\phi_2(x)$. Georgievskii published several works in which he developed a general theory \cite{Georgievskii1967ProceedingCoordMeetHydrotecbmologyNo25VNIIG-UnionSci-ResInstHydraulEng,GeorgievskiiMironichevaShulgin1968IzvAkadNaukUzbSSR,Georgievskii1971} for which modulating functions were a special case. Basovich applied Georgievskii's theory to estimating hydrological parameters in a stratum, where $\phi$ was composed of eigenfunctions of a fourth derivative operator combined with homogeneous Dirichlet boundary conditions \cite{Basovich1975FluidDyn}. Valeur pioneered the application of asymmetric $\phi$ functions adapted to fluorescence decay curves \cite{Valeur1978ChemicalPhysics}, e.g., $\phi(t)=t^\alpha(t-1)^\beta$, where $\alpha$, $\beta$ can be distinct.  Maletinsky created $\phi$'s via integrating Dirac delta distributions at subsets of the sample points, leading to a spline description of $\phi$ \cite{Maletinsky1979IFACProceedingsVolumes,Maletinsky1975IFACProceedingsVolumes}.

The 1980's and 1990's saw the development of more sophisticated $\phi$ functions, several of which have forms that were chosen specifically for their efficient computational implementations. For example, Pearson and Lee \cite{PearsonLee1985IEEETransAutomatContr,PearsonLee1985Control-TheoryAdvTechnol} based their $\phi$ functions on trigonometric functions of different frequencies and were the first to develop FFT-based computational speedups as well as provide advice on how many $\phi$'s to use. Kraus and Senning \cite{KrausSenning1987IFACProceedingsVolumes} developed a method to optimally choose $\phi$ based upon a Ritz-parameterized function family, while Jalai, Jordan, and Mackie demonstrated the value of aligning the peak of the test function with the peak of the measured cross-correlation function \cite{JalaliJordanMackie1992Automatica}. Pearson also showed how to deal with model terms that are not compatible with integration by parts (via using trigonometric $\phi$'s, spectral derivatives, and a binomial expansion)  \cite{Pearson1992Proc31stIEEEConfDecisControl}. Co and Ydstie built on the work of Pearson, demonstrating how it can be applied to several applications including reduced order modeling, multivariate system identification, and delay identifications \cite{CoYdstie1990ComputersChemicalEngineering}.
Patra and Unbenhauen \cite{PatraUnbehauen1995InternationalJournalofControl} adapted the Hartley Transform (enabling FFT-type efficient computations) to create $\phi$'s with a sum of trigonometric functions of different frequencies
\[
\phi(t)=\frac{1}{T_w}\sum_{j=0}^n (-1)^j \binom{n}{j}\textrm{cas} (2(n+k-j) \pi t/T_w)~,
\]
where $\textrm{cas}(t)=\cos(t)+\sin(t)$. There have also been efforts to use different classes of Wavelets as $\phi$ functions \cite{Schoenwald1993Proc32ndIEEEConfDecisControl,CarrierStephanopoulos1998AIChEJournal}. Lastly, Byrski and Fuksa then laid out a function space-based theory for optimal estimation (using modulating functions) for continuous SISO (Single Input Single Output) systems in $L^2$ \cite{ByrskiFuksa1995IntJModelSimul} as well as extended the MFM for simultaneous parameter and state estimation \cite{ByrskiFuksa1996IFACProcVol} (similar to the work 20 years later by Jouffroy and Reger \cite{JouffroyReger20152015IEEEConfControlApplCCA}).

In the first decade of the 21st century, there was not much research activity directly focused on the MFM.  There was, however, advances related to using Mikusi\'{n}ski’s operational calculus \cite{Mikusinski1987} to compute algebraic differentiators\footnote{For a good review, we direct the interested reader to \cite{OthmaneKiltzRudolph2022IntJSystSci}.} that could then be used in parameter (and state) estimation \cite{FliessSira-Ramirez2003ESAIMCOCV,MboupJoinFliess2009NumerAlgor,RegerJouffroy2009Proc48hIEEEConfDecisControlCDCHeldJointly200928thChinControlConf} as well as flatness-based motion planning \cite{RudolphWoittennek2008IntJControl}.\footnote{The term \emph{flatness} is from the control literature and is akin to the traditional concept of \emph{controllability}, but for nonlinear systems (as originally proposed in \cite{FliessLevineMartinEtAl1995IntJControl}).} These differentiators can be viewed as arising from a specific choice of $\phi$ in the MFM \cite{OthmaneRudolph202120215thIntConfControlFault-TolerSystSysTol}.  However, there is also evidence that clever choices of $\phi$ functions result in MFM-based parameter estimation methods which outperform these algebraic differentiators \cite{LiuGibaruPerruquetti2011201119thMediterrConfControlAutomMED}.

The other notable development in the 2000's related to the MFM arose in the context of fault detection.  Laroche et al., \cite{LarocheMartinRouchon2000IntJRobustNonlinearControl}, Lynch and Rudolph \cite{LynchRudolph2002InternationalJournalofControl}, and Rudolph and Woittennek \cite{RudolphWoittennek2008IntJControl} discovered that for a PDE model of fault detection, one could recast it as an optimal path problem. This path can be computationally solved for and allows for creation of an optimal modulating function (for this problem).

The 2010's and onward have seen a renewed and continuing interest in MFMs.  And, currently, the most prevalent forms of the $\phi$ functions used by researchers are asymmetric polynomials (e.g., see \cite{LiuGibaruPerruquetti2011201119thMediterrConfControlAutomMED,LiuLaleg-KiratiGibaruEtAl20132013AmControlConf,LiuLaleg-Kirati2015SignalProcessing,AldoghaitherLiuLaleg-Kirati2015SIAMJSciComput}) and Volterra linear integral operators \cite{PinLoveraAssaloneEtAl20132013AmControlConfa,PinAssaloneLoveraEtAl2015IEEETransAutomatContr,PinChenParisini2017Automatica,ChenPinNgEtAl2017IEEETransPowerElectron}.

There have been many excellent reviews over the years, some of which mention in passing the existence of weak form / MFMs \cite{PolisGoodson1976ProcIEEE}, while others focus almost exclusively on describing the different types of modulating functions \cite{PreisigRippin1993ComputChemEng,PreisigRippin1993ComputChemEnga,PreisigRippin1993ComputChemEngb,Pearson19954654NASAContractorReport}.  A good historical overview of the MFM can also be found in a recent dissertation \cite{Beier2023}.

We note that the overwhelming majority of the research on modulating functions is in the control theory and engineering literature, where a significant focus is placed on identifying and controlling ODE systems of the form $\sum_{i = 0}^n a_i(t)y^{(i)}(t) = \sum_{i = 0}^m b_i(t)u^{(i)}(t)$
(for observed variable $y$, control $u$, and parameters $a_i$ and $b_i$).  And, coinciding with the renewed interest in the MFM (starting in $\sim$2010), there is a significant body of recent work studying this class of ODE models  \cite{FedeleColuccio2010ApplMathComput,PinLoveraAssaloneEtAl20132013AmControlConfa,LiuLaleg-KiratiPerruquettiEtAl2014IFACProceedingsVolumes,JouffroyReger20152015IEEEConfControlApplCCA,PinAssaloneLoveraEtAl2015IEEETransAutomatContr,NoackRueda-EscobedoRegerEtAl20162016IEEE55thConfDecisControlCDC,PinChenParisini2017Automatica,ChenPinNgEtAl2017IEEETransPowerElectron,PinChenParisini2019Automatica,TianWeiLiuEtAl2019MechanicalSystemsandSignalProcessing}, with applications ranging from aerospace \cite{MorelliGrauer2020JAircr} to biomedical engineering \cite{AminFaghih2019IEEETransBiomedEng}.  However, weak form-based techniques have also been successfully used for spatio-temporal system identification as well.  Indeed, in their original paper, Loeb and Cahen proposed estimating the parameters in a diffusion equation \cite{LoebCahen1965Automatisme}. Perdreauville and Goodson extended Shinbrot's original work to apply to a beam equation and the Blasius equation with constant parameters, as well as to the diffusion equation with a spatially varying parameter \cite{PerdreauvilleGoodson1966JBasicEng}. Others like Fairman and Shen merged modulating function and method of lines (and denoted it as the \emph{Moment Functional Method}), applying it to the wave and diffusion equations with time varying coefficients \cite{FairmanShen1970InternationalJournalofControl}. More recent examples of applications of the MFM to PDEs include \cite{FischerDeutscher2016IFAC-PapersOnLine,AsiriLaleg-Kirati2017InverseProblSciEng,GhaffourNoackRegerEtAl2020IFAC-PapersOnLine}. In 2010, Janiczek noted that a generalized version of integration-by-parts could extend the MFM to work with fractional differential equations \cite{Janiczek2010BullPolAcadSciTechSci}. Accordingly, the MFM has found substantial success  in applications to fractional ODEs \cite{LiuLaleg-KiratiGibaruEtAl20132013AmControlConf,LiuLaleg-Kirati2015SignalProcessing,DaiWeiHuEtAl2016Neurocomputing,WeiLiuBoutat2017IEEETransAutomatContr,Gao2017IntJSystSci,BelkhatirLaleg-Kirati2018SystControlLett,WeiLiuBoutatEtAl2018SystControlLett,WeiLiuBoutat2019Automatica,LiuLiuBoutatEtAl2022CommunNonlinearSciNumerSimul} and fractional PDEs \cite{AldoghaitherLiuLaleg-Kirati2015SIAMJSciComput,AldoghaitherLaleg-Kirati2020JComputApplMath,GhaffourLaleg-Kirati2022SIAMJControlOptim}, including methods for estimating the order of differentiation \cite{AldoghaitherLiuLaleg-Kirati2015SIAMJSciComput,BelkhatirLaleg-Kirati2018SystControlLett,AldoghaitherLaleg-Kirati2020JComputApplMath}.


Broadly speaking, more general weak form methods have not seen widespread use due to 1) the challenge of selecting $\phi$ \cite{AsiriLiuLaleg-Kirati2021IEEEControlSystLett}, 2) a well-known statistical bias in EE-based inference \cite{Young1981Automatica,Regalia1994IEEETransSignalProcess}, and 3) the availability of software using output error (OE) methods. 
In an effort to address these issues, we recently proposed the Weak form Estimation of Nonlinear Dynamics (WENDy) parameter inference method,\footnote{\url{https://github.com/MathBioCU/WENDy}} providing an automated strategy for the creation of orthogonal $\phi$'s from multiresolution $C^\infty$ functions merged with a Generalized Least Squares (GLS) approach (to correct the statistical issues\footnote{Among the existing strategies for addressing this issue are 1) an Instrument Variables approach \cite{Young1970Automatica}, which yields a small bias for finite sample lengths, but at the loss of statistical efficiency \cite{Durbin1954RevIntStatInst} and 2) an iterative reweighting of the estimated covariance \cite{PearsonShen1993Proc32ndIEEEConfDecisControl}.}) \cite{BortzMessengerDukic2023BullMathBiol}.
The combination of these two techniques results in an algorithm that is notably fast, accurate, and noise-robust. Figure 2 depicts the relative errors vs.~walltime in using WENDy to estimate the parameters for the KS PDE from data with $20\%$ noise. In most cases, WENDy is notably more accurate and faster than using conventional OE methods, while being more accurate and only moderately slower than existing MFM methods (see \cite{BortzMessengerDukic2023BullMathBiol} for WENDy applied to ODE example systems).


\section{Coarse Graining}

\begin{figure}
\begin{center}
\begin{tabular}{@{}cc@{}}
 \includegraphics[clip,trim={0 0 0 0},height=0.4\textwidth]{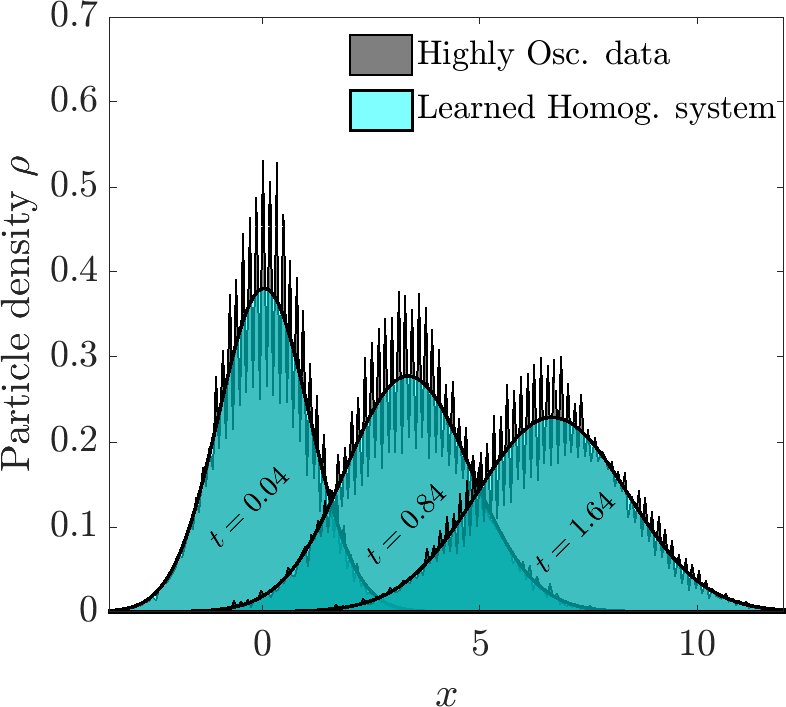} & 
 \includegraphics[clip,trim={0 0 0 0},height=0.4\textwidth]{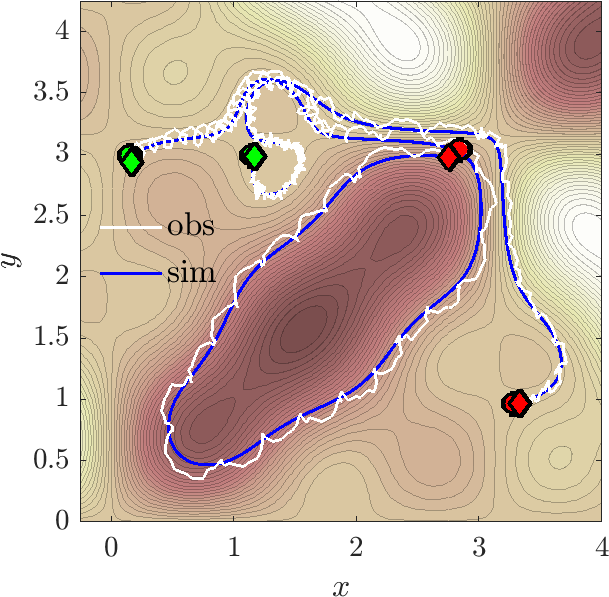}
 \end{tabular}
 \end{center}
 \caption{\label{coarse-grain}{\bf Coarse-graining}. 
 Left: Homogenization of a highly-oscillatory Fokker-Planck equation from particle data \cite{MessengerBortz2022PhysicaD}. Right: reduction of (noisy) coupled charged particle motion (white) to coarse-grained Hamiltonian dynamics (teal) including inference of background electric potential $\widehat{V}_\mathbf{E}$ (contours). Particles start at the green markers and end at the red ones. Note the proximity of the full dynamics (circles) to the coarse-grained model (diamonds) \cite{MessengerBurbyBortz2024SciRep}.
 }
\end{figure}

{\it Coarse graining} is the process of mapping a first-principles model to a lower order one characterized by effective descriptions of small-scale dynamics using larger-scale quantities of interest.
 Data-driven discovery of coarse-grained models is an active area of research, with recent works offering methods to discover hydrodynamic equations for active matter \cite{SupekarSongHastewellEtAl2023ProcNatlAcadSciUSA} and Fokker-Planck equations for random fields \cite{BakarjiTartakovsky2021JournalofComputationalPhysics}.
 %
 %
 %
In many cases, we derive a solution to the coarse-grained model as a limit of solutions to the first principles model (converging in a suitable weak topology). This naturally leads to questions about a role for weak form equation learning in coarse-graining applications. 
Indeed, for 1st-order stochastic IPS, WSINDy can discover the governing PDE corresponding to its mean-field McKean-Vlasov process from histograms of discrete-time samples of the IPS at the $N$-particle level \cite{MessengerBortz2022PhysicaD}.  Similarly, when observing diffusive transport with a highly-oscillatory spatially-varying diffusivity, WSINDy identifies the correct homogenized equation \cite{MessengerBortz2022PhysicaD}. Figure \ref{coarse-grain} (left) depicts histograms (shown in gray) collected from an $N$ particle system diffusing with a large (but finite) spatial frequency $\omega$, from which WSINDy is able to identify the correct $N\to \infty$, $\omega\to \infty$ homogenized system (the learned system is in teal).

Furthermore, for nearly-periodic Hamiltonian systems, WSINDy robustly identifies the correct leading-order Hamiltonian dynamics of reduced dimension obtained from averaging around an associated periodic flow that commutes with the full dynamics to leading order 
\cite{MessengerBurbyBortz2024SciRep}. 
In Figure \ref{coarse-grain} (right) noisy observations from an 8-dimensional coupled charged-particle system (in white), enable identification of a 4-dimensional coarse-grained Hamiltonian system (in blue), complete with accurate identification of the ambient background electric field $\widehat{V}_\mathbf{E}$. 

\section{Opportunities}
The purpose of this article is to draw attention to the successes and opportunities for research into weak form methods. Indeed, there are notable recent works suggesting that there are many advances to be discovered. For example, computationally, the sparse regression can be improved via the narrow-fit and trimming approach in WeakIdent \cite{TangLiaoKuskeEtAl2023JComputPhys} as well as by the optimal rank-1 updates in the Scalable Pruning for Rapid Identification of Null vecTors (SPRINT) algorithm
 \cite{Golden2024arXiv240509579}. Moreover, performing the bootstrap aggregation in Ensemble-SINDy (E-SINDy) using the weak form of the equations results in substantial reductions in model error \cite{FaselKutzBruntonEtAl2022ProcRSocMathPhysEngSci}. On the theoretical side, there is a novel proof of convergence (in a Reproducing Kernel Hilbert Space) of WSINDy-created surrogate models
\cite{RussoLaiu2024SIAMJApplDynSyst}. Moroever, combining an occupation kernel method \cite{RosenfeldRussoKamalapurkarEtAl2024SIAMJControlOptim} for equation learning with the weak form provides
an operator theoretic analog to weak-SINDy method \cite{RussoMessengerBortzEtAl202426thIntSympMathTheoryNetwSyst}. 
 A precise characterization of the class of models and type of noise for which WSINDy will always recover the true model (in the limit of continuum data) can be found here \cite{MessengerBortz2024IMAJNumerAnal}. 

Lastly, we note that all the advances described in this article are based on conventional techniques of statistics, applied analysis, and numerical analysis.  Applications of these techniques yield versions of equation learning, parameter inference, and model coarse graining that offer substantial robustness and accuracy and demonstrate that the weak form has broad applicability and utility beyond the well-known theoretical and computational methods.

\section{Acknowledgments}

This work is supported in part by the National Science Foundation grants 2054085 and 2109774, the National Institute of General Medical Sciences grant R35GM149335, the National Institute of Food and Agriculture grant 2019-67014-29919, and Department of Energy grant DE-SC0023346.

\section{Biographies}
Daniel Messenger is a Director’s Postdoctoral Fellow at Los Alamos National Laboratory. April Tran is a Rudy Horne Graduate Fellow in the Department of Applied Mathematics at the University of Colorado, Boulder. Vanja Dukic and David Bortz are professors in the Department of Applied Mathematics at the University of Colorado, Boulder.


\bibliography{SINEWSarXiv2024}
\bibliographystyle{siam}
\end{document}